\documentclass[lettersize,journal]{IEEEtran}
\usepackage{amsmath,amsfonts}
\usepackage{algorithm}
\usepackage{array}
\usepackage[caption=false,font=normalsize,labelfont=sf,textfont=sf]{subfig}
\usepackage{textcomp}
\usepackage{stfloats}
\usepackage{url}
\usepackage{verbatim}
\usepackage{graphicx}
\usepackage{cite}
\usepackage{algpseudocode}
\usepackage{amssymb}
\usepackage{placeins}
\usepackage{booktabs}
\usepackage[table]{xcolor}
\usepackage{multirow}

\begin{document}

\title{TransitReID: Transit OD Data Collection with Occlusion-Resistant Dynamic Passenger Re-Identification}

\author{Kaicong Huang$^1$, Talha Azfar$^1$, Jack M. Reilly$^1$, Ruimin Ke$^{1*}$,~\IEEEmembership{Senior Member, IEEE}
\thanks{$^{1}$ Civil and Environmental Engineering, Rensselaer Polytechnic Institute, 110 Eighth Street, Troy, NY USA 12180.}
\thanks{$^*$ Corresponding author.}
}

\markboth{Journal of \LaTeX\ Class Files,~Vol.~14, No.~8, August~2021}%
{Shell \MakeLowercase{\textit{et al.}}: A Sample Article Using IEEEtran.cls for IEEE Journals}


\maketitle

\begin{abstract}
Transit Origin-Destination (OD) data are fundamental for optimizing public transit services, yet current collection methods, such as manual surveys, Bluetooth/WiFi tracking, and Automated Passenger Counters, are often costly, device-dependent, or unable to support individual-level matching. Meanwhile, onboard surveillance cameras already deployed on most transit vehicles provide an underutilized opportunity for automated OD data collection. Leveraging this, we present TransitReID, a framework for individual-level and occlusion-resistant passenger re-identification (ReID) tailored to transit environments. TransitReID introduces three key components: (1) an occlusion- and viewpoint-robust ReID algorithm that integrates a variational autoencoder-guided region-attention mechanism with selective feature pooling to emphasize visible and discriminative body regions; (2) a Hierarchical Storage and Dynamic Matching (HSDM) mechanism that adapts static ReID matching to dynamic bus operations while balancing accuracy, memory, and speed; and (3) a multi-threaded edge implementation that enables near real-time OD estimation while preserving privacy through local data processing. We also construct a new Transit ReID dataset with over 17,000 images captured from real bus front/rear cameras under diverse occlusion and viewpoint conditions. Experimental results show that TransitReID achieves state-of-the-art ReID performance, attaining 88.3\% R-1 accuracy on the proposed transit ReID dataset and sustaining 80--90\% OD estimation accuracy in both simulations and real-world operation, with deployment supported on NVIDIA Jetson edge devices. This work provides an algorithmic and system-level foundation for scalable, privacy-preserving automated transit OD collection.
\end{abstract}


\section{Introduction}
Transit Origin-Destination (OD) data collection plays a critical role in the fields of transportation engineering for urban planning, service frequency adjustments, and route optimization \cite{rong2024interdisciplinary}. For public transportation systems, OD data represents passenger demand and plays a crucial role in designing routes more efficiently to minimize travel time, reduce operating costs, and optimize driver scheduling.  

\begin{figure}
    \centering
    \includegraphics[width=1.0\columnwidth]{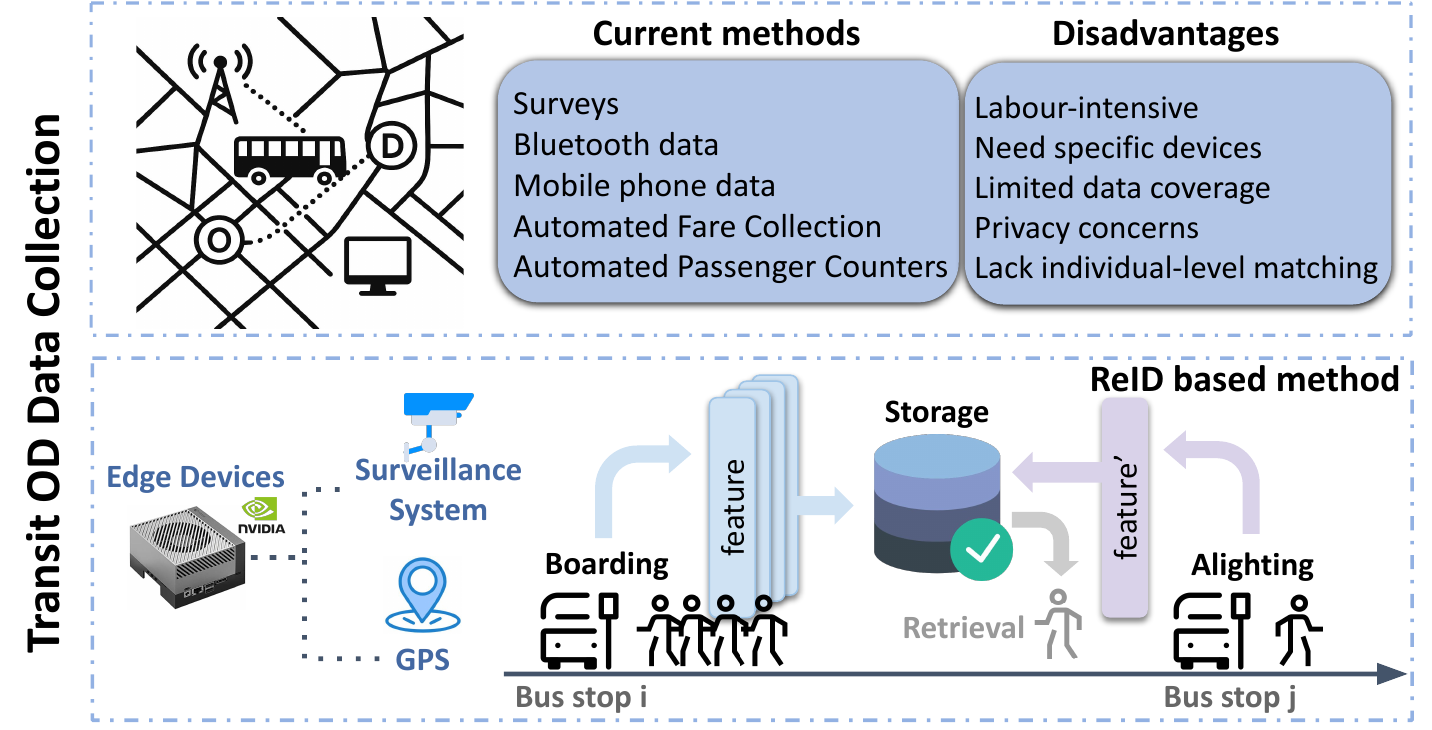}
    \caption{Traditional methods for estimating transit OD data compared with the proposed method. Our approach utilizes onboard camera data to achieve seamless and accurate passenger re-identification and collect individual-level OD data.}
    \label{intro}
\end{figure}

As shown in Fig. \ref{intro}, traditional methods for estimating transit OD data rely on manual approaches such as surveys, which are labor-intensive and suffer from low response rates \cite{bernardin2017understanding}. More advanced methods, such as those utilizing mobile phone data\cite{larijani2015investigating} and Bluetooth technology \cite{ozbay2017real}, require passengers to carry specific devices with WiFi or Bluetooth enabled, which limits coverage and might raise privacy concerns due to the unique identifiers of the device. Automated Passenger Counters (APCs) \cite{ji2015transit} also assist in estimating OD data, however, they can only capture the counts and locations of boarding and alighting passengers and fail to match passengers at the individual level.
From a transit planner's perspective, individual-level OD information goes beyond aggregate flows by preserving each passenger's boarding-alighting relationship, enabling finer demand forecasting, user-segment analysis, and detection of irregular behaviors such as fare evasion or abnormal trajectories.

To address these needs, video-based re-identification (ReID) techniques offer a promising solution. Unlike device-dependent approaches, video-based ReID captures continuous passenger trajectories without requiring user cooperation and, when combined with anonymization strategies, can achieve this while mitigating privacy concerns. Meanwhile, most transit vehicles in the US are already equipped with onboard cameras for liability and surveillance purposes, presenting a valuable opportunity to repurpose them as smart sensors for video-based transit OD data collection \cite{shimada2019person}. By integrating these cameras with ReID techniques, edge computing devices, and GPS, seamless individual-level matching and OD data collection can be achieved.

Despite the advantages of camera-based methods over existing techniques, their adoption in transit OD data collection remains limited due to several challenges, particularly occlusion. For example, during peak times, passengers may obscure each other from the camera view, leading to information loss and background confusion. Recent research on ReID has increasingly emphasized occlusion, with existing approaches broadly falling into two categories: local attention–based methods and feature recovery–based methods. The former constructs attention mechanisms by focusing on different local areas to suppress interference from occluded parts \cite{ning2024occluded}, while the latter attempts to recover information from occluded areas by associating temporal or spatial information~\cite{10314802, 10038705}. Although some methods achieve strong performance on general ReID benchmarks, they still struggle with dynamic viewpoints and complex occlusion patterns, and often fail to transfer effectively to real-world application scenarios such as transit vehicles. A key limitation is that they usually force the model to implicitly discover informative regions or reconstruct missing cues, rather than explicitly evaluating the reliability and contribution of each body region under arbitrary occlusion conditions.

Additionally, unlike traditional ReID studies, our work focuses on the problem of transit OD data collection, hence the gallery is dynamic, changing with the variations in passengers boarding and alighting at each station. Consequently, transit ReID evolves from a problem of matching 1 query to $N$ options, into a dynamic 1 to \((N_1, N_2, ..., N_K)\) problem, where \(N_k\) represents the size of the gallery updated at station $k \in K$.

The limited computational resources of edge devices present an additional challenge in executing complex analytical algorithms \cite{liu2022bringing, azfar2024deep}. Fast and accurate estimation of OD data is crucial for minimizing latency in passenger flow analysis. To address this, some efforts have focused on model lightweighting techniques, such as knowledge distillation and pruning, or on optimizing neural network architectures specifically for edge hardware \cite{azfar2023incorporating}. Nevertheless, lightweight models often face a trade-off between efficiency and accuracy.

\begin{figure*}
\centering
\includegraphics[width=\textwidth]{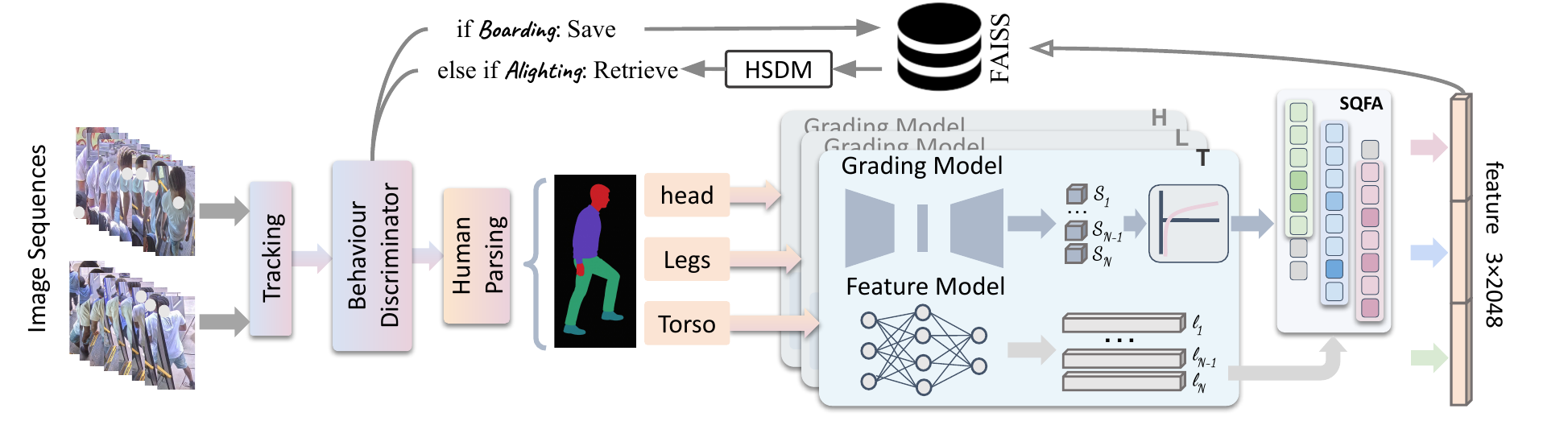}
\caption{
The framework of the proposed TransitReID. The surveillance system captures image sequences of the passengers boarding and alighting at different stops. 
After behavior detection, each image is processed by a human parsing network, which divides the image into three body parts: `head', `torso', and `legs'. Each part is then fed into a grading model and a feature generation model to obtain a quality score and an initial feature representation. Finally, a selective quality feature averaging (SQFA) unit integrates these outputs into the final feature vector. At boarding, the feature is first stored in the gallery; at alighting, the system employs the hierarchical storage and dynamic matching (HSDM) mechanism to identify the corresponding feature from the gallery.} 
\label{framework}
\end{figure*}

To jointly tackle the aforementioned issues, we propose TransitReID, an individual-level transit OD data collection method that incorporates occlusion-resistant and dynamic passenger ReID. The overall framework is shown in Fig. \ref{framework}, featuring two principal technical innovations:
(1) An occlusion-robust ReID algorithm employing a region-attention module to adaptively weight less-obscured body segments with clearer viewpoints, guided by a co-trained Variational Autoencoder (VAE) grading model that explicitly and robustly assesses body-part integrity under complex occlusions and dynamic viewpoints. This is followed by a Selective Quality Feature Averaging (SQFA) strategy that preserves discriminative information through concatenated weighted averages of region embeddings.
(2) A Hierarchical Storage and Dynamic Matching (HSDM) mechanism specifically designed for dynamic transit OD matching, which incorporates Hot/Cold Storage and a Snatch mechanism to balance accuracy, storage, and speed.
A multi-threaded design is implemented to support near real-time operation on transit vehicles equipped with edge devices, without requiring any modifications to the models or architectures. Notably, the framework processes video locally on edge devices and stores only high-level, non-identifiable features, ensuring a privacy-preserving and non-intrusive solution.

Another issue is that current video ReID datasets 
are not specifically designed for occlusion scenarios, while most occluded person ReID datasets 
are image-based and not suited for training tasks under transit surveillance video conditions. 
Additionally, person images in those datasets are captured from consistent angles, whereas in transit scenarios, the angles of passenger images captured by surveillance cameras change continuously. This variability is crucial for training a robust and generalizable model under transit bus operational scenarios. Therefore, we construct a transit ReID dataset, collected from real bus operational settings, including 17,636 images of 157 passengers captured by cameras at the front and rear doors.


The major contributions of this work are as follows:
\begin{itemize} 
\item We propose TransitReID to enable individual-level transit OD data collection, employing an attention-based passenger ReID algorithm that explicitly evaluates part-level quality and contribution for robust matching under variable occlusions and viewpoint changes.

\item A hierarchical storage and dynamic matching mechanism tailored for actual transit scenarios is designed, transforming the traditional static ReID problem into a dynamic ReID challenge. 

\item Introduce an efficient multi-threaded edge implementation for real-time smart transit applications, validated through both bus route simulations and real-world operation.
\end{itemize}

\section{Related Work}
\subsection{Transit OD Data Collection}
Transit OD data collection refers to the systematic process of gathering information on the origins and destinations of individual trips within a public transportation network. 
Passenger-level transit OD data are commonly estimated through surveys \cite{baltes2002customer, Zalewski2019PublicTR}, but these methods are labor-intensive and often suffer from low response rates. To reduce manual effort, prior studies have explored automated data sources, including Automated Fare Collection (AFC)-based OD and load profile estimation under round-trip assumptions \cite{jafari2021procedure}, smart-card trajectory search for daily travel reconstruction \cite{li2011estimating}, Automated Passenger Counter (APC)-based optimization for network-level OD approximation \cite{liu2021optimization}, and Radio Frequency Identification (RFID)-based Boarding-In and Boarding-Out (BIBO) systems for passenger monitoring \cite{gonzalez2020detailed}. Other studies have used mobile phone data \cite{larijani2015investigating}, Bluetooth technology \cite{ozbay2017real}, and multi-source data fusion, such as integrating Geographic Information System (GIS) data with passenger Integrated Circuit (IC) card data for OD matrix construction \cite{kong2021research}. However, these approaches still face limitations, including high deployment and maintenance costs \cite{bernardin2017understanding}, dependence on passengers carrying specific devices, limited data coverage, and potential privacy risks in large-scale public transportation applications \cite{zhao2004planning}.

While most existing methods capture only aggregate counts and locations, they lack the precision required to track and match individual passengers, which is essential for detailed behavioral analysis. Video-based passenger re-identification with transit surveillance cameras provides a potential solution. In \cite{shimada2019person}, YOLO was used for passenger detection, followed by feature extraction and vector-distance matching, but the adverse effects of occlusion were not considered. To mitigate occlusion, some studies \cite{guo2021bus, zhao2022detection} focused on passenger head images inside buses. However, head-only methods discard informative body regions, have limited discriminative power for passengers with similar headgear, and are sensitive to lighting variations. Our proposed methodology not only maximizes the useful information extracted from each passenger through an attention mechanism but also enhances privacy protection by processing data locally on onboard edge devices, ensuring that only high-level features are stored.

\subsection{Occluded Re-Identification}
Occlusion is a major challenge in vision-based person ReID, and existing methods mainly focus on body-part feature learning and visible region localization \cite{ning2024occluded}. CNN-based methods, such as the region-based quality predictor in \cite{song2018region}, learn weighted regional features, but their fixed regions are less suitable for transit scenarios with continuously changing camera-passenger distances and viewpoints. Pose-guided methods use human landmarks or posture-guided attention to locate reliable visible regions \cite{miao2019pose, gao2020pose, wang2022pose}, while recent parsing-guided approaches exploit human parsing for part attention and fine-grained local representation learning \cite{chen2024part}. Although these methods are related to our motivation, their performance depends on accurate pose or parsing estimation, which can be unreliable under severe crowding, low resolution, and complex occlusions in transit environments. Therefore, instead of relying solely on part localization, a more general and robust part-level quality assessment strategy is needed, which motivates the proposed VAE-based grading model and selective feature pooling strategy.
Other methods include feature recovery\cite{10314802, 10038705, hou2024feature}, graph convolution\cite{wang2020high}, and multi-model methods\cite{farooq2022axm, hafner2022cross}. However, most existing methods are designed for static ReID benchmarks and may not generalize well to transit scenarios with dynamic viewpoints, illumination variations, and complex occlusions. Moreover, the gallery in a transit vehicle is inherently dynamic as passengers board and alight. Therefore, we design a hierarchical storage and dynamic matching (HSDM) mechanism specifically adapted for dynamic ReID.

\section{Methodology}
\subsection{Problem Definition}
We consider a transit bus equipped with two surveillance cameras, namely a front-door camera \(C_f\) and a rear-door camera \(C_r\). The gallery set \(G = G_f \cup G_r\) contains all onboard passenger images captured by both cameras:
\begin{equation}
G_f = \{g_f^1,\ldots,g_f^{M_f}\},
\quad
G_r = \{g_r^1,\ldots,g_r^{M_r}\}
\end{equation}
where \(M_f\) and \(M_r\) denote the numbers of gallery images from the front and rear cameras, respectively.

At bus stop \(s\), the query set \(Q^s= Q_f^s \cup Q_r^s\) consists of alighting passenger images from both doors:
\begin{equation}
Q_f^s = \{q_f^{s,1},\ldots,q_f^{s,N_f^s}\},
\quad
Q_r^s = \{q_r^{s,1},\ldots,q_r^{s,N_r^s}\}
\end{equation}
where \(N_f^s\) and \(N_r^s\) are the numbers of alighting passengers captured at the front and rear doors.

Given a feature extractor \(\phi(\cdot)\), each passenger image \(x\) is represented as:
\begin{equation}
\phi(x) \in \mathbb{R}^{d}
\end{equation}
For each query image \(q_i^s \in Q^s\), passenger re-identification is formulated as nearest-neighbor retrieval from the gallery:
\begin{equation}
\hat{g}_i
=
\arg\min_{g_j \in G}
D\left(\phi(q_i^s), \phi(g_j)\right)
\end{equation}
where \(D(\cdot,\cdot)\) is the feature distance metric and \(\hat{g}_i\) denotes the matched onboard passenger image.

\begin{figure}
\centering
\includegraphics[width=0.8\linewidth]{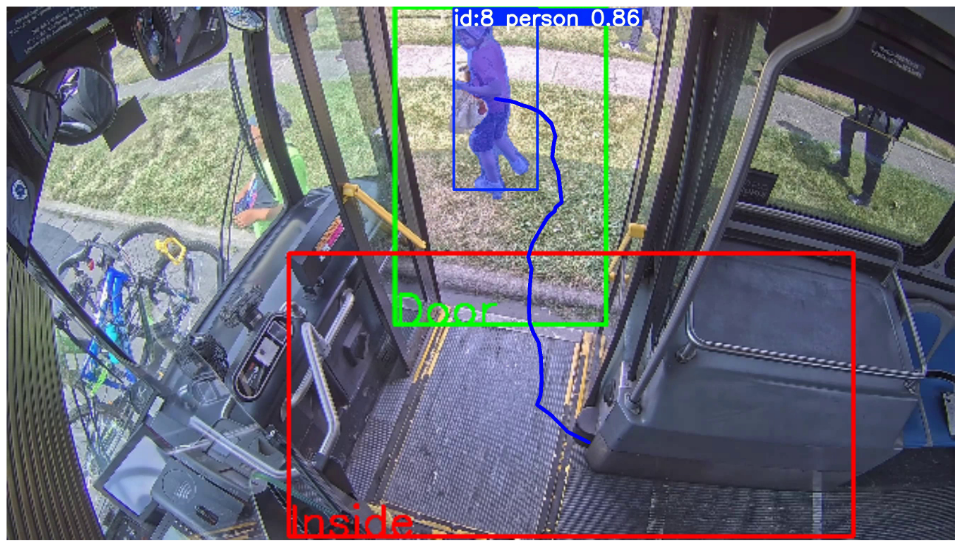}
\caption{One alighting example. The green box represents the outdoor area and the red box represents the area inside the bus. The blue trajectory shows the moving of the passenger.}
\label{alighting}
\end{figure}

\subsection{Detection and Tracking}
This work uses surveillance videos recorded by the front- and rear-door cameras of the bus. Boarding and alighting passengers are detected using YOLO11 accelerated with NVIDIA TensorRT. Following our previous work \cite{ye2025trbposter}, two image regions, \(R_{door}\) and \(R_{inside}\), are predefined to represent the outside and inside areas of the bus door, respectively. Passenger behavior is then identified by a rule-based classifier according to the spatial relationship between passenger bounding-box centers and these regions of interest (ROIs), as illustrated by the alighting example in Fig. \ref{alighting}.

Define the starting point as \( S = (x_{s}, y_{s}) \) and ending point as \( E = (x_{e}, y_{e}) \) for a trajectory, there are four possible conditions:
\begin{equation}
\begin{aligned}
    &\big(S \in R_{\text{inside}} \big) \land \big(S \notin R_{\text{door}} \big) \land \big(E \in R_{\text{door}} \big) \\
    &\big(S \in R_{\text{door}} \big) \land \big(E \in R_{\text{inside}} \big) \land \big(E \notin R_{\text{door}} \big) \\
    &\big(S \in R_{\text{inside}} \big) \land \big(S \notin R_{\text{door}} \big) \land \big(E \in R_{\text{inside}} \big) \land \big(E \notin R_{\text{door}} \big) \\
    &\big(S \in R_{\text{door}} \big) \land \big(E \in R_{\text{door}} \big)
\end{aligned}
\end{equation}
These equations correspond to trajectories of \textit{1) alighting, 2) boarding, 3) moving inside the bus, and 4) moving while remaining outside}. 
This processing effectively filters out false boarding and alighting behaviors, such as when a passenger lingers near the door but ultimately does not board the bus. Only trajectories satisfying the first two cases are retained, with their corresponding locations recorded as key-value pairs.

\subsection{Data Preprocessing}
Occlusion in transit vehicles severely degrades conventional ReID performance, especially during peak boarding and alighting periods. Target passengers may be partially blocked by seats, luggage, or nearby passengers \cite{wang2022feature}, causing missing body cues and introducing irrelevant regions into the bounding box, which leads to feature confusion. To reduce such interference, we apply body segmentation and use local body-part representations instead of relying only on global features. Following \cite{li2020self}, each passenger is segmented into seven parts using the Pascal-Person-Part dataset \cite{chen2014detect}, which are further merged into three semantic regions: \textbf{Head}, \textbf{Torso} including torso and arms, and \textbf{Legs} including upper and lower legs.

\subsection{Grading Model}
\begin{figure}
    \centering
    \includegraphics[width=\linewidth]{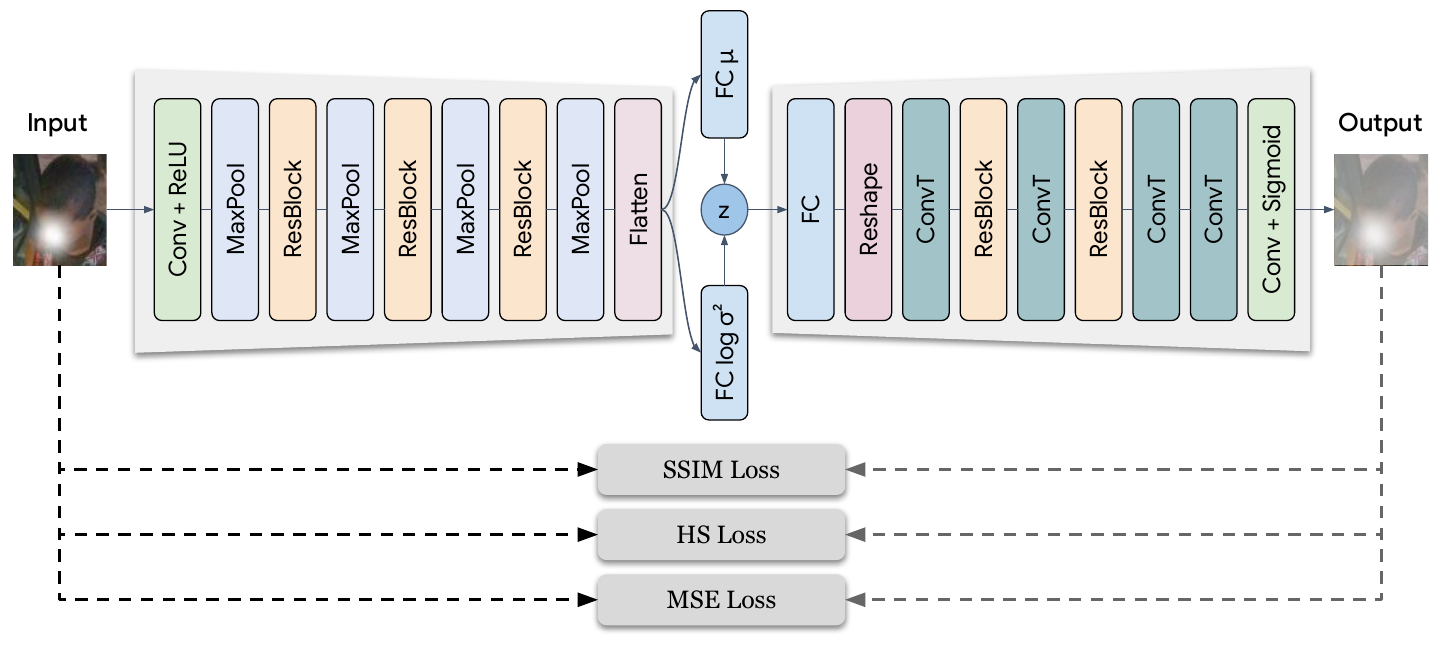}
    \caption{Structure of the proposed grading model. The reconstruction module employs a Variational Autoencoder (VAE), and its output is further compared with the raw image for score calculation.}
    \label{VAE}
\end{figure}

We design a Variational Autoencoder (VAE)-based grading model to assign quality scores to images of different body parts, where a higher score indicates less occlusion and a clearer viewpoint. As shown in Fig. \ref{VAE}, the grading model is jointly trained with the feature extraction network, and the resulting score is used to calculate image weights in subsequent stages. The underlying logic is straightforward: through training, the VAE is encouraged to better reconstruct high-quality body-part images while poorly reconstructing those with severe occlusions or unfavorable viewpoints. Since the score is derived from the holistic reconstruction consistency between the input and reconstructed output, the model can adapt to complex occlusion patterns and dynamic viewpoints regardless of how occlusion occurs or from which angle the body part is observed. Before scoring, each parsed body-part image is resized to \(128 \times 128\), converted to grayscale, and its zero-valued background pixels are replaced by a small constant value. The raw reconstruction score is defined as:
\begin{equation}
S = \sigma_1 \text{SSIM}(X, \hat{X})
  + \sigma_2 \text{HS}(X, \hat{X})
  + \sigma_3 S_{\text{MSE}}(X, \hat{X})
\end{equation}
where $X$ and $\hat{X}$ denote the original and reconstructed images. $\sigma_1$, $\sigma_2$, and $\sigma_3$ are their weights.

The Structural Similarity Index Measure (SSIM) \cite{wang2004image} measures structural and perceptual similarity. The Histogram Similarity (HS) is computed as the cosine similarity between normalized 256-bin intensity histograms of the original and reconstructed images. To reduce the influence of the artificial background value, the corresponding histogram bin is suppressed before computing histogram similarity.

The mean squared reconstruction error is converted into a similarity score:
\begin{equation}
S_{\text{MSE}} =
\frac{1}{1 + \exp(k(\text{MSE} - m))}
\end{equation}
where $k$ and $m$ are hyperparameters. In our implementation, $k=50$ and $m=0.03$.

\subsubsection*{\bf Score Mapping}
A logarithmic transformation is applied to normalize the original scores to \([0,1]\). By compressing the score distribution, this transformation strongly suppresses low-quality images with severe occlusion while largely preserving the contribution of high-quality images with minor occlusion.


Finally, the combined score $S$ is clipped and mapped through the logarithmic normalization:
\begin{equation}
I_S =
\frac{\log(1 + \alpha(S - S_{\min}))}
{\log(1 + \alpha(S_{\max} - S_{\min}))}
\end{equation}
where $S_{\min}$, $S_{\max}$, and $\alpha$ are hyperparameters. We set $S_{\min}=0.08$, $S_{\max}=0.50$, and $\alpha=20$.

\subsection{Selective Quality Feature Averaging}
Given an image sequence of a passenger extracted from video, \(\{I_1, I_2, \ldots, I_l\}\), where \(l\) is the sequence length, each image is fed into a feature extraction network to obtain a fixed-length feature vector. We adopt GRL \cite{liu2021watching} as the backbone, which captures both global and local feature correlations across video frames. The three body parts, namely Head, Torso, and Legs \((H,T,L)\), are processed separately to generate the corresponding feature sets:
\begin{equation}
\mathcal{F} = \{ F_k = \{ f_1^k, f_2^k, \dots, f_l^k \} \}_{k \in \{H, T, L\}}
\end{equation}

The score sets output by the grading model are denoted by:
\begin{equation}
\mathcal{S} = \{ S_k = \{ s_1^k, s_2^k, \dots, s_l^k \} \}_{k \in \{H, T, L\}}
\end{equation}
We use \(l = 8\) frames per person for consistency.

\begin{figure}
\centering
\includegraphics[width=\linewidth]{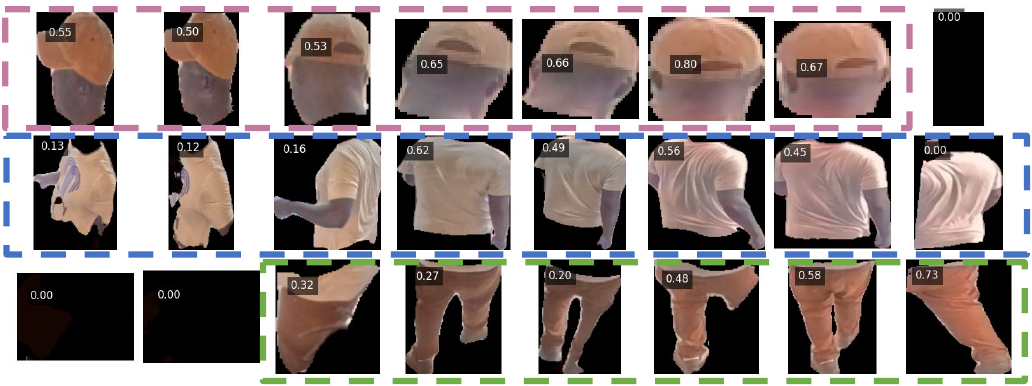}
\caption{Selective quality feature averaging (SQFA). Each image is attached with a score reflecting its degree of occlusion. Averaging only frames with scores above the threshold (as shown in the boxes) would preserve the features of high quality.}
\label{Averaging}
\end{figure}

Unlike the baseline model \cite{liu2021watching}, we do not aggregate all images in a sequence, since low-quality frames may dilute discriminative cues and cause confusion among passengers with similar local body features. Instead, only images whose scores exceed a predefined threshold are retained, preserving high-quality contributions to the final representation, as shown in Fig. \ref{Averaging}. The passenger-level feature vector is then computed as:
\begin{equation}
\widetilde{F}
=
\operatorname{Concat}_{k \in \{H,T,L\}}
\left(
\frac{1}{L_k}
\sum_{i \in \mathcal{V}_k}
S_k^i F_k^i
\right)
\end{equation}
where \(k \in \{H,T,L\}\) denotes the Head, Torso, and Legs regions, respectively. \(F_k^i\) and \(S_k^i\) are the feature vector and quality score of the \(i\)-th image for body part \(k\). \(\mathcal{V}_k\) is the set of valid images whose scores exceed the predefined threshold, and \(L_k = |\mathcal{V}_k|\) is the number of retained images. The resulting features are stored in Facebook AI Similarity Search (FAISS) \cite{douze2024faiss}.

\subsection{Hierarchical Storage and Dynamic Matching}
\begin{figure}
\centering
\includegraphics[width=\linewidth]{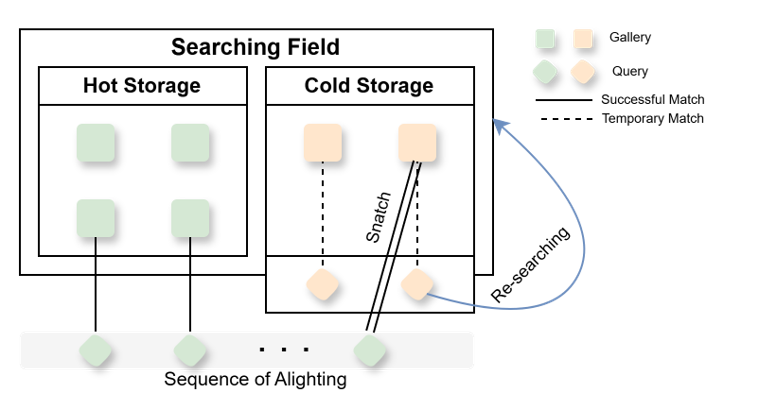}
\caption{Structure of hierarchical storage and dynamic matching (HSDM).}
\label{HSDM}
\end{figure}

Unlike conventional ReID tasks, transit-vehicle scenarios are dynamic and spatiotemporally continuous, leading to a continuously updated onboard gallery. Removing matched passengers can reduce the search space, lower subsequent matching errors, and save memory on edge devices, but direct deletion is risky because imperfect matching may cause error accumulation. Therefore, we propose a hierarchical storage and dynamic matching (HSDM) method for practical transit ReID, as shown in Fig. \ref{HSDM}.

The HSDM consists of three major steps as follows, and the processing logic is summarized in Algorithm \ref{alg:HSDM}.

\subsubsection{\bf Hot/Cold Matching and Storage}
Passengers are processed sequentially by bus stop from \(1\) to \(S\). At each stop, the features of boarding passengers are inserted into the FAISS index to update the gallery. For each alighting passenger, we retrieve the Top-\(K\) candidates from FAISS, where \(K=5\) in this work, and obtain their gallery IDs and distances. The matching confidence is defined by the distance gap between the rank-1 and rank-2 candidates:
\begin{equation}
\gamma = 1 - \frac{d_{\text{rank-1}}}{d_{\text{rank-2}}}
\end{equation}
where \(d_{\text{rank-1}}\) and \(d_{\text{rank-2}}\) denote the distances of the first and second nearest candidates, respectively. A confidence threshold \(\sigma\) is determined from the distance distributions of positive and negative pairs. If \(\gamma > \sigma\), the match is treated as reliable and assigned to Hot Storage, where the matched gallery ID is removed from the FAISS index. Otherwise, the match is assigned to Cold Storage: the rank-1 ID is kept as a temporary match, while the remaining candidates from rank-2 to rank-\(K\) are stored as alternatives without deleting any gallery entry.

\subsubsection{\bf Snatch}
If a new alighting passenger \(q_i\) matches a gallery ID \(g_j\) already held by another temporary passenger \(q_o\), the ID ownership is reassigned according to confidence:
\begin{equation}
\operatorname{owner}(g_j)=
\arg\max_{q \in \{q_i,q_o\}} \gamma(q,g_j)
\end{equation}
When \(\gamma(q_i,g_j)>\gamma(q_o,g_j)\), \(q_i\) snatches \(g_j\), and the match is either confirmed and removed from FAISS if \(\gamma(q_i,g_j)>\sigma\), or kept in Cold Storage otherwise. The original holder \(q_o\) then selects the next available candidate from its stored Top-\(K\) alternatives for rematching.

\subsubsection{\bf Final Update}
After all stations have been processed, passengers who have not been matched and are still in a temporary matching state automatically transition to a successful match, and the matching accuracy for each station is then calculated.

\begin{figure*}
    \centering
    \includegraphics[angle=270, width=1\textwidth]{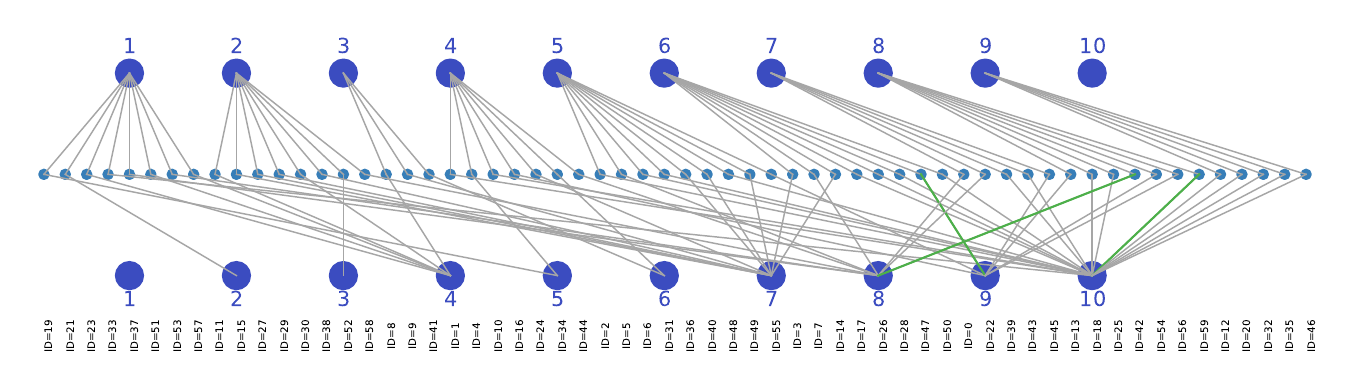}
    \caption{OD data collection in 10 bus stops. The first and last rows represent the 10 boarding and alighting stops, while the middle row corresponds to 60 passengers. Gray lines indicate correct matches, whereas green lines represent incorrect matches.}
    \label{ODlines}
\end{figure*}

\begin{figure*}
    \centering
    \includegraphics[angle=0, width=1\textwidth]{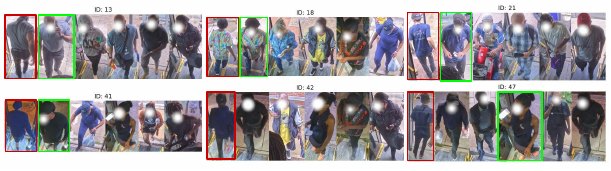}
    \caption{Top 5 retrieved results in the simulation with 10 bus stops. The target passengers getting off/on the bus are marked with red/green boxes. From left to right are the top 1 to 5 matched results. The IDs 13, 18, 21, 41 are correctly matched, while the IDs 42 and 47 are incorrect.}
    \label{final_result_mask}
\end{figure*}

\begin{algorithm}[t]
\caption{HSDM: Overall Process}
\label{alg:HSDM}
\setlength{\baselineskip}{10pt}
\begin{algorithmic}[1]
\State Initialize FAISS index, Hot Storage, and Cold Storage
\For{$i=1$ to $S$}
    \State Add features of boarding passengers at stop $i$ to FAISS
    \For{each alighting passenger $p$ at stop $i$}
        \State Retrieve Top-$K$ candidates $\mathcal{C}_p=\{(g_k,d_k)\}_{k=1}^{K}$ from FAISS
        \State Compute confidence $\gamma_p = 1 - d_1/d_2$
        \If{$g_1$ is temporarily held by another passenger $p'$ in Cold Storage}
            \If{$\gamma_p > \gamma_{p'}$}
                \State Reassign $g_1$ from $p'$ to $p$; let $p'$ rematch using its remaining candidates
            \Else
                \State Let $p$ rematch using candidates $\{g_2,\ldots,g_K\}$
            \EndIf
        \EndIf
        \If{$\gamma_p > \sigma$}
            \State Assign $g_1$ to $p$ as a reliable match and remove $g_1$ from FAISS
        \Else
            \State Store $g_1$ as the temporary match of $p$ and keep $\{g_2,\ldots,g_K\}$ as alternatives
        \EndIf
    \EndFor
\EndFor
\For{each passenger $p$ remaining in Cold Storage}
    \State Confirm its temporary match as the final match
\EndFor
\State Compute final matching accuracy
\end{algorithmic}
\end{algorithm}

\section{Experiments}
\subsection{Implementations}
We evaluate the performance of this work from two aspects. First, we evaluate the passenger ReID on our own dataset and compare it with other state-of-the-art approaches. The training is on a single RTX4090 24GB GPU with the batch size, epochs, and initial learning rate set to 8, 60, and 0.0001 with a decay rate 0.1 after 30 epochs, respectively. Next, we run the simulation for transit OD data collection along a manually configured bus route on an NVIDIA Jetson AGX Orin.
The confidence threshold of HSDM is set to 0.12 based on the confidence distribution of the training data. This value maximizes the discrimination between positive and negative samples. The weights $\sigma_1, \sigma_2, \sigma_3 $ of the joint similarity are set to $0.2, 0.6, 0.2$, respectively. To speed up processing, only 8 images are randomly sampled to form a passenger's tracklet. The loss function follows the same setting in \cite{liu2021watching}. Performance metrics include R-1, 5, 10, 20 and mean Average Precision (mAP)

\subsection{Bus ReID Dataset}
To address the lack of occluded re-identification datasets in public transportation scenarios, we construct a new dataset from several hours of real bus surveillance footage captured by front- and rear-door cameras. The dataset contains over 17,000 images of 157 passengers, including 97 passengers for training and 60 for validation. It covers diverse occlusion patterns, such as passenger-passenger and passenger-object occlusions, as well as variations in viewpoint and illumination. These characteristics provide challenging conditions for model training and help evaluate generalization in complex transit environments.

\begin{table}[t]
\centering
\caption{Comparison with related methods.}
\label{tab:bus_reid2025}
\renewcommand{\arraystretch}{1.15}
\setlength{\tabcolsep}{5.5pt}
\resizebox{\linewidth}{!}{
\begin{tabular}{l c c c c c c}
\toprule
\textbf{Method} & \textbf{Reference} & \textbf{R-1} & \textbf{R-5} & \textbf{R-10} & \textbf{R-20} & \textbf{mAP} \\
\midrule
PGFA~\cite{miao2019pose}        & ICCV'19 & 60.9 & 70.2 & 75.3 & 80.4 & 44.3 \\
HORelD~\cite{wang2020cvpr}      & CVPR'20 & 69.0 & 75.6 & 78.2 & 82.0 & 58.8 \\
GRL~\cite{liu2021watching}      & CVPR'21 & 75.0 & 96.7 & \textbf{100.0} & \textbf{100.0} & 83.6 \\
TransReID~\cite{He_2021_ICCV}   & ICCV'21 & 83.6 & 87.1 & 89.1 & 91.4 & 78.8 \\
PFD~\cite{wang2022pose}         & AAAI'22 & 78.8 & 84.0 & 86.6 & 89.3 & 74.6 \\
CLIP-ReID~\cite{li2023clip}     & AAAI'23 & 55.4 & 68.4 & 75.0 & 82.8 & 45.7 \\
TF-CLIP~\cite{yu2024tf}         & AAAI'24 & 86.7 & 96.7 & 98.3 & \textbf{100.0} & 90.7 \\
ReIDMamba~\cite{gu2026reidmamba}& TMM'25  & 74.6 & 79.2 & 81.9 & 84.0 & 64.5 \\
\midrule
\rowcolor{gray!15}
\textbf{TransitReID} & \textbf{Ours} & \textbf{88.3} & \textbf{98.3} & \textbf{100.0} & \textbf{100.0} & \textbf{92.6} \\
\bottomrule
\end{tabular}
}
\end{table}

\subsection{Evaluation on ReID Performance}
Table \ref{tab:bus_reid2025} reports the performance of TransitReID on traditional ReID tasks compared with several state-of-the-art methods. TransitReID achieves the best overall performance, improving R-1 accuracy by 1.6\% and mAP by 1.9\%. Although the baseline GRL is not specifically designed for occlusion, its video-based temporal modeling enables it to outperform some occlusion-oriented methods. Building upon it, TransitReID further exploits clear-view body regions and suppresses occlusion interference, leading to clear gains in R-1 and mAP. The similar R-5 results between baseline and TransitReID indicate that baseline can often retrieve the correct identity within the top candidates, while our method improves hard-case discrimination by promoting the correct match to rank-1 under multi-view and complex occlusion conditions.

\subsubsection*{\bf Quality Information Extraction}
\begin{figure}
    \centering
    \includegraphics[width=\linewidth]{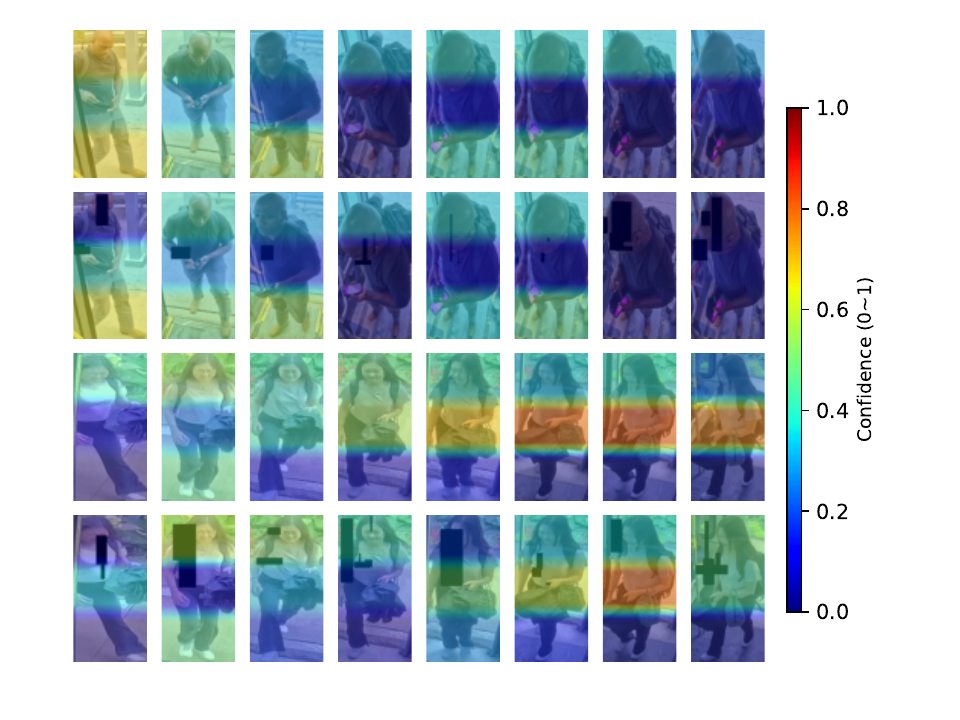}
    \caption{Score visualization after applying the grading model. Each image is divided into three segments, corresponding to three body parts. The score is mapped to colors from dark blue to red, where a higher score indicates higher quality.}
    \label{score_vis}
\end{figure}

Fig. \ref{score_vis} visualizes four tracklet groups, where each image is divided into three body-part segments for score visualization only and may not exactly match real body boundaries. In the third row, the leg region obtains a lower score due to obstruction by an umbrella, while the exposed head and torso receive higher scores. In the first row, the head score becomes higher than the torso and leg scores from the fourth image onward, even without external occlusion, because the increasing camera-view angle compresses visible body information and causes self-occlusion. The second and fourth rows contain manually added occlusions, whose affected regions receive lower scores than the original images, demonstrating the model's sensitivity to occlusion. These results indicate that the proposed scoring strategy effectively captures both external occlusion and viewpoint-induced quality degradation.

\subsection{Evaluation on Dynamic Transit OD Data Collection}
We randomly assign the 60 validation passengers to 10 bus stops and generate their boarding and alighting behaviors. The bus is empty before the first stop, where passengers can only board, while all remaining onboard passengers must alight at the terminal stop. At each intermediate stop, boarding is processed before alighting. Although this differs from real-world cases where the two behaviors may occur simultaneously, it overestimates the gallery size at each stop, thereby increasing matching difficulty and providing a stricter evaluation of the proposed method.

Fig. \ref{ODlines} presents the OD collection results, where the first and last rows denote the 10 boarding and alighting stops, and the middle row represents the 60 passengers. Prediction is performed between passengers and alighting stops, with gray and green lines indicating correct and incorrect matches, respectively. Among 60 passengers, only three are incorrectly matched, yielding an OD collection accuracy of 95\%. This metric differs from the ReID retrieval accuracy reported in the ablation study, as it evaluates the correctness of the final inferred alighting stops rather than retrieval ranking performance.

Fig. \ref{final_result_mask} shows representative retrieval results under different viewpoints, passenger occlusions, and daytime/nighttime conditions, where red and green boxes mark boarding and alighting images of the target passenger, respectively, and the results are ordered from top-1 to top-5. IDs 13, 18, 21, and 41 are correctly matched, demonstrating the ability of our method to extract discriminative features in complex transit environments. IDs 42 and 47 illustrate failure cases: ID 47 alights earlier and incorrectly matches the ground truth of ID 42 as its top-1 result, making this gallery sample unavailable when ID 42 later alights. This shows that although HSDM reduces error propagation, its performance still depends on the confidence threshold and cannot guarantee perfect matching. Nevertheless, the 95\% OD accuracy validates the effectiveness of the proposed approach for transit data collection.

\begin{table}[t]
\centering
\caption{Statistics of each stop's information and processing time.}
\label{tab:TIME}
\renewcommand{\arraystretch}{1.15}
\setlength{\tabcolsep}{6pt}
\resizebox{\linewidth}{!}{
\begin{tabular}{lccccc}
\toprule
\rowcolor{gray!12}
\textbf{Stop ID} & \textbf{1} & \textbf{2} & \textbf{3} & \textbf{4} & \textbf{5} \\
\midrule
Board/Alight & 8/0 & 8/1 & 3/1 & 7/6 & 9/2 \\
$T_F$        & 33.4 & 30.3 & 13.2 & 42.8 & 36.0 \\
$T_M$        & -- & $1.55{\times}10^{-3}$ & $1.38{\times}10^{-3}$ & $6.30{\times}10^{-3}$ & $2.81{\times}10^{-3}$ \\
$T_{\mathrm{mean}/P}$ & -- & 3.37 & 3.30 & 3.31 & 3.00 \\
\midrule
\rowcolor{gray!12}
\textbf{Stop ID} & \textbf{6} & \textbf{7} & \textbf{8} & \textbf{9} & \textbf{10} \\
\midrule
Board/Alight & 8/2 & 5/11 & 7/9 & 5/7 & 0/21 \\
$T_F$        & 32.7 & 52.8 & 52.0 & 39.3 & 68.2 \\
$T_M$        & $2.72{\times}10^{-3}$ & $1.22{\times}10^{-2}$ & $1.00{\times}10^{-2}$ & $7.92{\times}10^{-3}$ & $2.14{\times}10^{-2}$ \\
$T_{\mathrm{mean}/P}$ & 3.27 & 3.30 & 3.06 & 2.81 & 3.25 \\
\bottomrule
\end{tabular}
}
\end{table}

\subsection{Inference Time Analysis on Edge Devices}
We evaluate inference efficiency on an NVIDIA Jetson AGX Orin to simulate edge deployment in realistic transit scenarios, with results shown in Table \ref{tab:TIME}. Here, Board/Alight denotes the number of passengers boarding or alighting at each stop, \(T_F\) is the total feature extraction time, and \(T_M\) is the ReID retrieval time. The average processing time is about 3 seconds per passenger, mainly limited by edge-device computation. To support practical deployment, we implement a multi-threaded framework for real-time smart transit applications.

\begin{figure}
    \centering
    \includegraphics[width=0.9\linewidth]{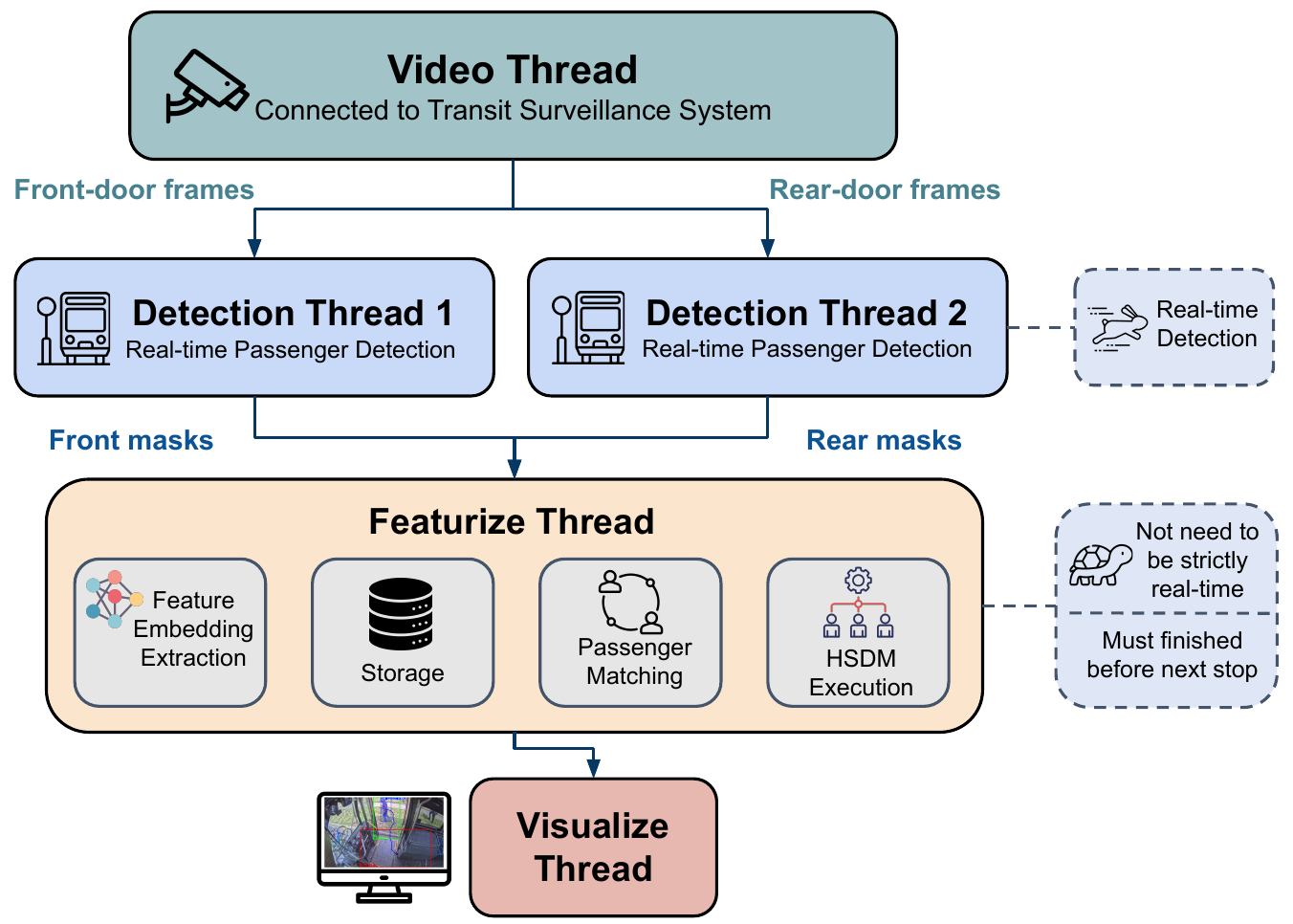}
    \caption{Multi-threaded edge implementation.}
    \label{thread}
\end{figure}

As shown in Fig. \ref{thread}, the video thread captures frames from the front and rear door cameras and feeds them into two detection threads, where YOLO enables real-time passenger detection. The featurize thread receives detection masks, extracts embeddings, and performs storage, matching, and HSDM execution, while the visualization thread is used only for debugging. Since detection runs in real time, the system can operate smoothly; feature extraction and matching only need to finish before the bus reaches the next stop. For example, if 10 passengers board at stop \(i\), the processing should be completed within about 30 seconds before arrival at stop \(i+1\).

\begin{figure*}
    \centering
    \includegraphics[width=\linewidth]{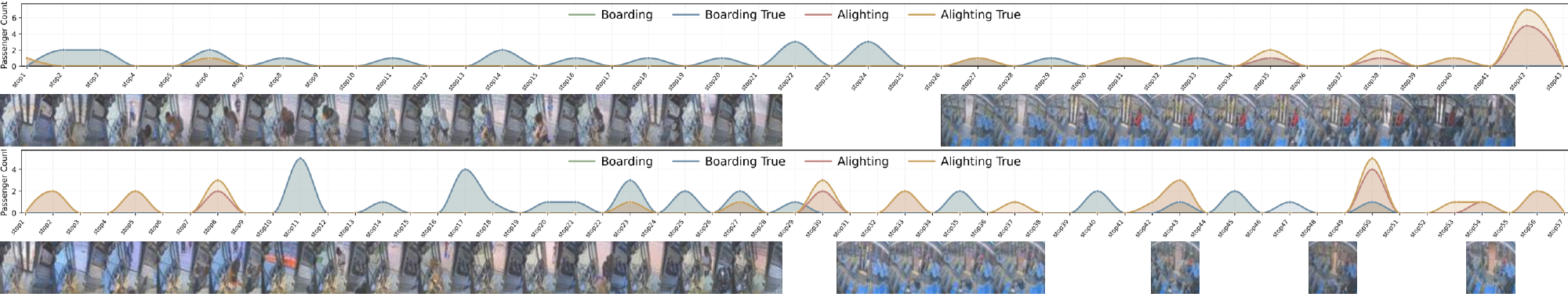}
    \caption{Evaluation results on real bus route operation data. The upper and lower panels correspond to the 7--8 AM and 20--21 PM peak-hour videos, respectively. White video clips indicate periods with no available video signal.}
    \label{real}
\end{figure*}

\begin{table}[t]
    \centering
    \caption{Accuracy of tracking and OD estimation on real operation.}
    \label{tab:real}
    \begin{tabular}{lcccc}
        \toprule
        \textbf{Category} & \textbf{Estimated Count} & \textbf{True Count} & \textbf{Accuracy} & \textbf{OD Acc} \\
        \midrule
        \rowcolor{gray!15}
        \multicolumn{5}{c}{\textbf{7--8 AM}} \\
        \midrule
        Boarding  & 26 & 26 & 100.00\% & \multirow{2}{*}{81.8\%} \\
        Alighting & 13 & 17 & 76.47\%  &  \\
        \midrule
        \rowcolor{gray!15}
        \multicolumn{5}{c}{\textbf{20--21 PM}} \\
        \midrule
        Boarding  & 29 & 30 & 96.67\% & \multirow{2}{*}{88.2\%} \\
        Alighting & 23 & 28 & 82.14\% &  \\
        \bottomrule
    \end{tabular}
\end{table}

\subsection{Evaluation on Real Bus Route}
In collaboration with a transit agency, we collected real-world bus operation data during two peak-hour periods: 7--8 AM with 43 stops and 20--21 PM with 57 stops, as shown in Fig. \ref{real}. The inputs are continuous videos captured by the front- and rear-door cameras throughout the operation periods. As reported in Table \ref{tab:real}, the boarding detection accuracy is consistently high, while the alighting accuracy is relatively lower. Manual inspection shows that most alighting errors occur when passengers stand too close to each other, making it difficult for the tracking pipeline to separate individual passengers. The OD estimation accuracy is consistent with the simulation results, demonstrating the effectiveness of the proposed system in real bus operation scenarios.

\subsection{Ablation Study}
\subsubsection*{\bf HSDM}
The effectiveness of HSDM is evaluated by simulations with 5 to 15 bus stops, where each setting is repeated 20 times and the average matching rate is reported in Table \ref{tab:HSDM}. Without HSDM, the R-1 rate increases only slightly as the number of stops grows, mainly because fewer gallery samples remain at each stop, reducing matching difficulty. With HSDM, the R-1 rate consistently improves across all settings, showing that the proposed mechanism can better handle low-confidence matches by keeping alternative candidates for later correction.
Moreover, HSDM maintains stable R-1 performance under different numbers of stops. This is because Cold Storage makes the gallery depend not only on the number of onboard passengers but also on matching confidence, allowing uncertain samples to remain available instead of being prematurely removed. In contrast, R-5 and R-10 show limited changes with or without HSDM, indicating that the base ReID model already retrieves most simple cases within the top candidates, while HSDM mainly improves difficult rank-1 decisions.

\subsubsection*{\bf Contributions from Different Components}
We evaluate the contribution of different components on the proposed bus ReID dataset, as shown in Table \ref{tab:component}. First, we remove one of the three body parts individually, and observe that both mAP and R-1 decrease compared to using the full body. “Grad” denotes the grading model, which brings improvements of 5.2\% in mAP and 10\% in R-1. SQFA slightly improves R-1 by 0.2\% and R-5 by 3.3\%.

\begin{table*}[t]
\centering
\caption{Performance across different numbers of bus stops. Each cell reports results without / with HSDM.}
\label{tab:HSDM}
\renewcommand{\arraystretch}{1.15}
\setlength{\tabcolsep}{4.2pt}
\resizebox{\linewidth}{!}{
\begin{tabular}{lccccccccccc}
\toprule
\rowcolor{gray!15}
\textbf{Stop ID} 
& \textbf{5} & \textbf{6} & \textbf{7} & \textbf{8} & \textbf{9} & \textbf{10} 
& \textbf{11} & \textbf{12} & \textbf{13} & \textbf{14} & \textbf{15} \\
\midrule
\textbf{R-1}  
& 0.87 / 0.90 & 0.87 / 0.90 & 0.87 / 0.91 & 0.88 / 0.91 & 0.86 / 0.90 
& 0.88 / 0.92 & 0.89 / 0.92 & 0.88 / 0.91 & 0.90 / 0.92 & 0.89 / 0.90 & 0.89 / 0.90 \\

\textbf{R-5}  
& 0.96 / 0.94 & 0.95 / 0.94 & 0.95 / 0.96 & 0.95 / 0.96 & 0.95 / 0.94 
& 0.95 / 0.95 & 0.95 / 0.96 & 0.95 / 0.95 & 0.95 / 0.96 & 0.95 / 0.94 & 0.95 / 0.94 \\

\textbf{R-10} 
& 0.96 / 0.94 & 0.95 / 0.94 & 0.95 / 0.96 & 0.95 / 0.96 & 0.95 / 0.94 
& 0.95 / 0.95 & 0.95 / 0.96 & 0.95 / 0.95 & 0.95 / 0.96 & 0.95 / 0.94 & 0.95 / 0.94 \\
\bottomrule
\end{tabular}
}
\end{table*}

\begin{table}[t]
\centering
\caption{Ablation study on the effectiveness of each component on the Bus ReID dataset.}
\label{tab:component}
\renewcommand{\arraystretch}{1.15}
\setlength{\tabcolsep}{5.5pt}
\resizebox{\linewidth}{!}{
\begin{tabular}{ccccc ccc}
\toprule
\multicolumn{5}{c}{\textbf{Components}} 
& \multicolumn{3}{c}{\textbf{Performance}} \\
\cmidrule(lr){1-5} \cmidrule(lr){6-8}
\textbf{Head} & \textbf{Legs} & \textbf{Torso} & \textbf{Grad} & \textbf{SQFA} 
& \textbf{mAP} & \textbf{R-1} & \textbf{R-5} \\
\midrule
-- & \checkmark & \checkmark & \checkmark & \checkmark 
& 84.8 & 75.0 & 96.7 \\
\checkmark & -- & \checkmark & \checkmark & \checkmark 
& 88.1 & 80.0 & \textbf{98.3} \\
\checkmark & \checkmark & -- & \checkmark & \checkmark 
& 90.0 & 85.0 & 95.0 \\
\checkmark & \checkmark & \checkmark & -- & \checkmark 
& 89.9 & 83.3 & 96.7 \\
\checkmark & \checkmark & \checkmark & -- & -- 
& 87.2 & 78.3 & 95.0 \\
\checkmark & \checkmark & \checkmark & \checkmark & -- 
& 92.4 & \textbf{88.3} & 95.0 \\
\checkmark & \checkmark & \checkmark & \checkmark & \checkmark 
& \textbf{92.6} & \textbf{88.3} & \textbf{98.3} \\
\bottomrule
\end{tabular}
}
\end{table}

\section{Conclusion}
This paper presents TransitReID, a framework for individual-level and dynamic transit OD data collection. TransitReID addresses two key challenges in transit environments: severe occlusion and continuous operation. For occlusion handling, it introduces a region-aware ReID mechanism with a VAE-based grading model to weight visible body regions and fuse discriminative features. For continuous operation, it develops a Hierarchical Storage and Dynamic Matching (HSDM) strategy with Hot/Cold Storage and Snatch mechanisms to balance accuracy, storage, and retrieval efficiency. A multi-threaded design is further implemented to support near real-time operation on edge devices, while local feature processing helps reduce privacy risks. Experiments demonstrate the effectiveness of the proposed framework, achieving 88.3\% R-1 accuracy on the proposed transit ReID dataset and an average OD estimation accuracy of 80--90\% in both simulation and real world operation. 

Beyond its technical contributions, TransitReID demonstrates the potential of integrating edge AI, computer vision, and intelligent transportation infrastructure into practical cyber-physical systems. The proposed framework offers a scalable and privacy-aware approach for continuous passenger sensing, which can support transit operators in understanding mobility patterns, optimizing service planning, and developing data-driven intelligent transportation applications.


\bibliographystyle{IEEEtran}
\bibliography{ref}

\begin{IEEEbiography}[{\includegraphics[width=1in,height=1.25in,clip,keepaspectratio]{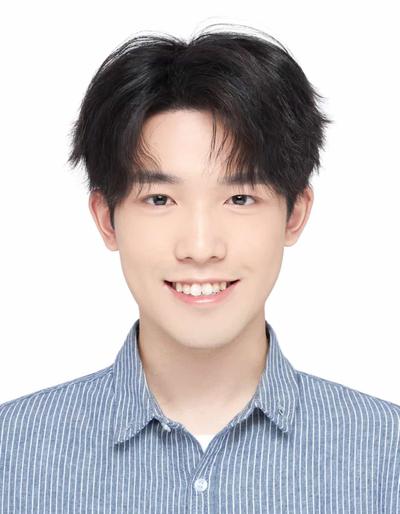}}]{Kaicong Huang}
is a Ph.D. student in Transportation Engineering at Rensselaer Polytechnic Institute (RPI) since August 2024. He holds a M.S. in Robotics from National University of Singapore (2024), and a B.E. in Electronic Engineering from Sun Yat-sen University (2023). His research interests include 3D vision, robotics and autonomous vehicles, and intelligent transportation systems.
\end{IEEEbiography}

\vspace{-10mm}

\begin{IEEEbiography}[{\includegraphics[width=1in,height=1.25in,clip,keepaspectratio]{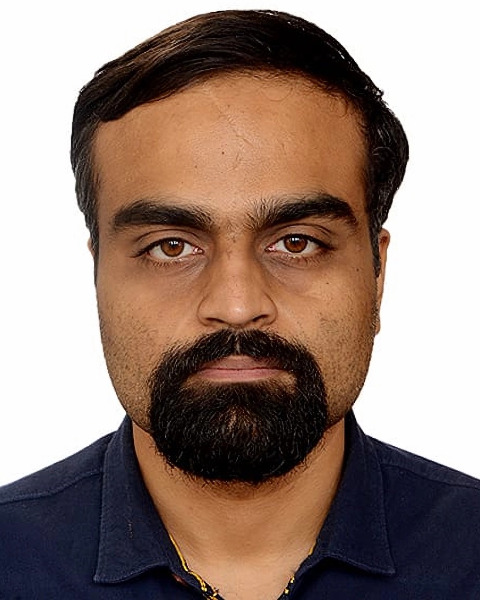}}]{Talha Azfar}
is a Ph.D. student in Transportation Engineering at Rensselaer Polytechnic Institute (RPI) since August 2023. He began his Ph.D. in the University of Texas at El Paso in 2022. His research focus lies in the intersection of computer vision, digital twins, edge computing, and intelligent transportation infrastructure. Talha holds a Masters degree in Systems Engineering from PIEAS, Islamabad (2016), and a Bachelors in Electrical Engineering from NUST, Islamabad (2014).
\end{IEEEbiography}

\vspace{-10mm}

\begin{IEEEbiography}[{\includegraphics[width=1in,height=1.25in,clip,keepaspectratio]{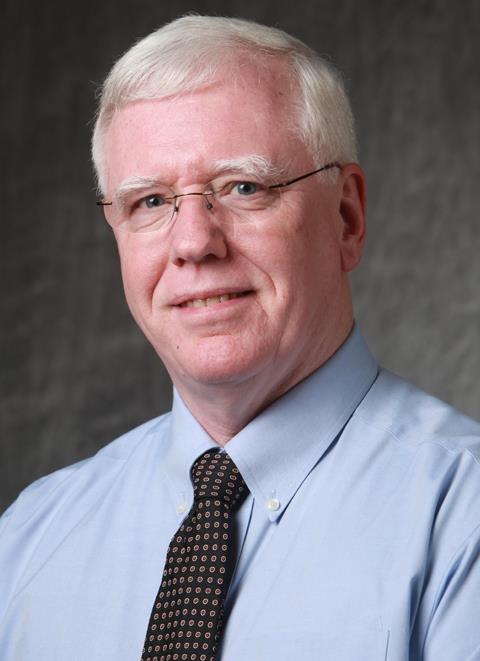}}]{Jack M. Reilly}
is a Professor of Practice in the Department of Civil and Environmental Engineering at Rensselaer Polytechnic Institute in Troy, NY where he teaches courses in transportation engineering, policy and design. He has been on the RPI faculty for 15 years. Dr. Reilly is a proud RPI alumnus having received a bachelor’s, master’s and doctoral degree from that school. During most of his professional career, he was on the management staff of the Capital District Transportation Authority, responsible for service planning, capital project development, information technology and federal relations. Dr. Reilly was the project manager for the Rensselaer Rail Station. He has done extensive consulting work in transportation planning and operations throughout the United States, and has served as a consultant to the World Bank on Intelligent Transportation System projects in India and China.
\end{IEEEbiography}

\vspace{-10mm}

\begin{IEEEbiography}[{\includegraphics[width=1in,height=1.25in,clip,keepaspectratio]{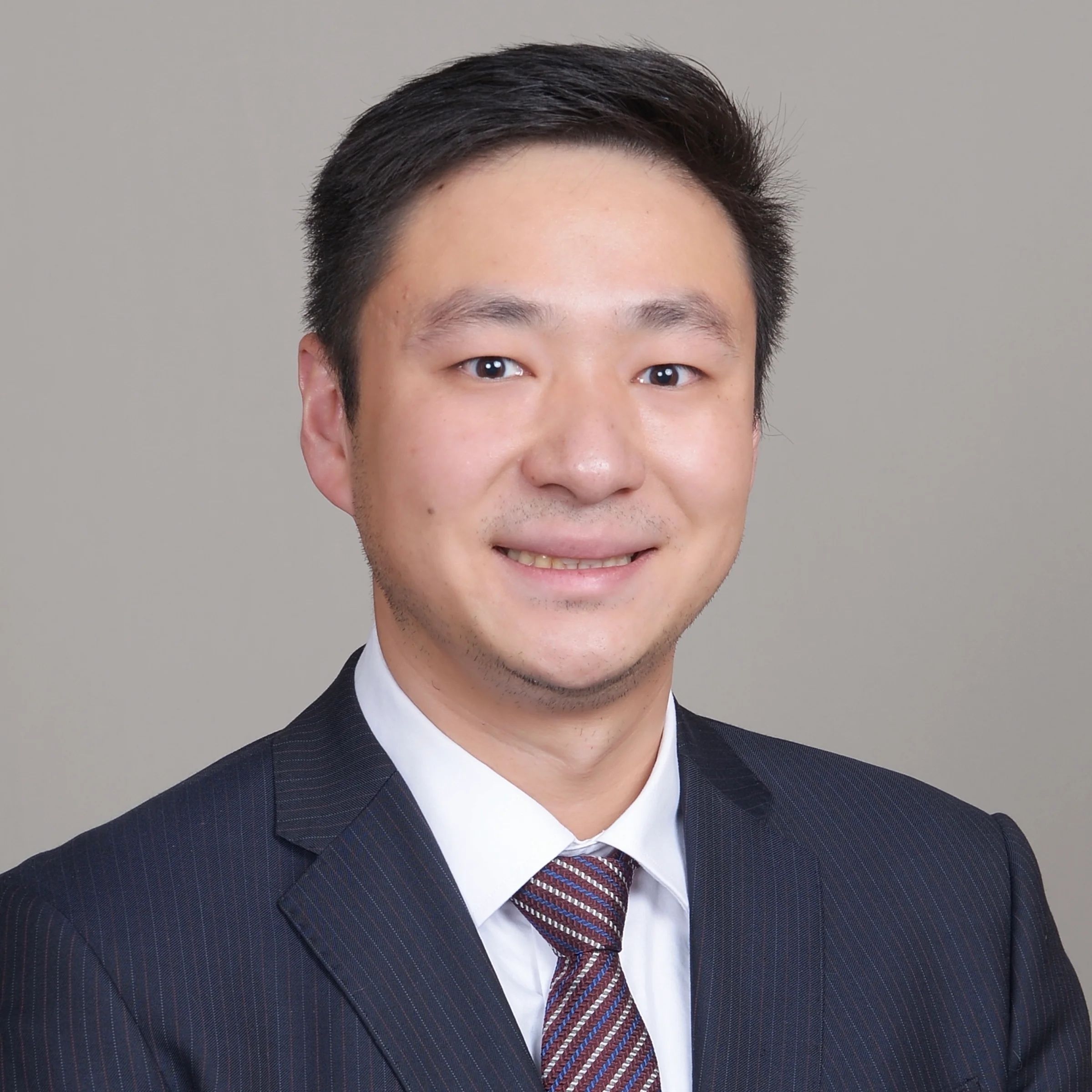}}]{Ruimin Ke} (Senior Member, IEEE) is an Assistant Professor of Transportation Engineering at Rensselaer Polytechnic Institute. Dr. Ke possesses profound expertise in intelligent transportation systems (ITS), with a specialization in machine learning algorithms and advanced computing systems tailored for ITS applications. Dr. Ke's academic background includes a Ph.D. and M.S. in Transportation Engineering from the University of Washington, an M.S. in Computer Science from the University of Illinois Urbana-Champaign, and a B.E. in Automation from Tsinghua University. 
\end{IEEEbiography}

\vfill

\end{document}